\documentclass[11pt,a4paper]{article}
\usepackage[hyperref]{emnlp2018}
\usepackage{times}
\usepackage{latexsym}
\usepackage{booktabs}
\usepackage{multirow}

\usepackage{paralist}
\usepackage{subcaption}
\usepackage{graphicx}
\usepackage{url}

\usepackage{amsmath}
\usepackage{amsfonts}

\aclfinalcopy % Uncomment this line for the final submission
\def\aclpaperid{61} %  Enter the acl Paper ID here

\title{Measuring Issue Ownership using Word Embeddings
\Thanks{This research was supported by the Swedish Research Council under contract 2017-02429}}

\author{Amaru Cuba Gyllensten \\
  RISE AI \\
  {\tt amaru.cuba.gyllensten@ri.se} \\\And
  Magnus Sahlgren \\
  RISE AI \\
  {\tt magnus.sahlgren@ri.se} \\}

\date{}

\begin{document}

\maketitle

\begin{abstract}
Sentiment and topic analysis are common methods used for social media monitoring. Essentially, these methods answers questions such as, ``what is being talked about, regarding \emph{X}", and ``what do people feel, regarding \emph{X}". In this paper, we investigate another venue for social media monitoring, namely {\em issue ownership} and {\em agenda setting}, which are concepts from political science that have been used to explain voter choice and electoral outcomes. 
% In short, the theory states that voters value certain issues, and cast votes according to the party which they feel best address these issues. Media has a 
We argue that issue alignment and agenda setting can be seen as a kind of semantic source similarity of the kind ``how similar is source \emph{A} to issue owner \emph{P}, when talking about issue \emph{X}", and as such can be measured using word/document embedding techniques. We present work in progress towards measuring that kind of conditioned similarity, and introduce a new notion of similarity for predictive embeddings. We then test this method by measuring the similarity between politically aligned media and political parties, conditioned on bloc-specific issues. 

\end{abstract}

\section{Introduction}
{\em Social Media Monitoring} (SMM; i.e.~monitoring of online discussions in social media) has become an established application domain with a large body of scientific literature, and considerable commercial interest. The subfields of {\em Topic Detection and Tracking} \cite{allan1998topic,sridhar2015unsupervised} and {\em Sentiment Analysis} \cite{Turney:2002:TUT:1073083.1073153,Pang:2008:OMS:1454711.1454712,Liu:2012:SAO:3019323,Pozzi:2016:SAS:3044756} are both scientific topics spawned entirely within the SMM domain. In its most basic form, SMM entails nothing more than counting occurrences of terms in data; producing frequency lists of commonly used vocabulary, and matching of term sets related to various topics and sentiments. More sophisticated approaches use various forms of probabilistic topic detection (such as Latent Dirichlet Allocation) and sentiment analysis based on supervised machine learning. 

The central questions SMM seeks to answer are {\em ``what do users talk about?''} and {\em``how do they feel about it?''}. Answers to these questions may provide useful insight for market research and communications departments. It is apparent how product and service companies may use such analysis to gain an understanding of their target audience. It is also apparent how such analysis may be used in the context of elections for providing an indication of citizens' opinions as manifested in what they write in social media. There are numerous studies attempting to use various forms of social media monitoring techniques to predict the outcome of elections, with varying success \cite{bermingham2011using,ceron2015using}.

Most notably, the recent examples of the inadequacy of standard opinion measuring techniques to forecast the most recent US election and the Brexit demonstrate that for certain questions related to measuring mass opinion, standard SMM techniques may be inadequate. Political scientists have used the concepts of {\em agenda setting} and {\em issue ownership} to explain voter choice and election outcomes \cite{kluver,IJoC3008,Stubager2018}. In short, the issue ownership theory of voting states that voters identify the most credible party proponent of a particular issue and cast their ballots for that issue owner \cite{BELANGER2008477}. Agenda setting refers to the media's role in influencing the importance of issues in the public agenda \cite{mccombs}. Note that current social media monitoring techniques are unable to measure these concepts in a satisfactory manner; it does not suffice to measure the occurrence of certain keywords, since most parties tend to use the same vocabulary to discuss issues, and sentiment analysis does not touch upon the issue ownership and agenda setting questions. What is needed for measuring issue ownership and agenda setting is a way to measure {\em language use}, i.e.~when talking about an issue, to which extent does the language used align with issue owner \emph{A} vs.~issue owner \emph{B}. 

We argue that issue alignment can be seen as a kind of semantic source similarity of the kind ``how similar is source \emph{A} to issue owner \emph{P}, when talking about issue \emph{X}", and as such can be measured using word/document embedding techniques. To measure that kind of conditioned similarity we introduce a new notion of similarity for predictive word embeddings. This method enables us to manipulate the similarity measure by weighting the set of entities we account for in the predictive scoring function. The proposed method is applied to measure similarity between party programs and various subsets of online text sources, conditioned on bloc specific issues.
The results indicate that this conditioning disentangles similarity. We can, for example, observe that while the Left Party representation is, overall, similar to that of nativist media, it differs significantly on nativist issue, while this effect is not seen to the same extent on more mainstream left wing or right wing media.

\section{Vector Similarity}

Vector similarity has been a foundational concept in natural language processing ever sine the introduction of the {\em vector space model} for information retrieval by \citet{Salton:1971:SRS:1102022}. In this model, queries and document are represented as vectors in term space, and similarity is expressed using cosine similarity. 
% formel kanske?
The main reason for using cosine similarity in the vector space model is that it normalizes for vector length; the fact that a document (or query) contains a certain word is more important than how many times it occurs in the document.

The vector space model was the main source of inspiration for early work on {\em vector semantics}, such as Latent Semantic Analysis \cite{deerwester,landauer_solution_1997} and the works on {\em word space models} by \citet{Schutze:1992:DM:147877.148132,Schütze93wordspace}. These works continued to embrace cosine similarity as the similarity metric of choice, since length normalization is equally desired when words are represented by vectors whose elements encode (some function of) co-occurrences with other words. Contemporary research on {\em distributional semantics} \cite{sahlgren2006word,Bullinaria2007,Turney:2010:FMV:1861751.1861756,pennington2014glove} still use largely the same mathematical machinery as the vector space model, and cosine similarity is still the preferred similarity metric due to its simplicity and use of length normalization. Even neural language models, which originate from the neural network community, employ cosine similarity to quantify similarity between learned representations \cite{NIPS2013_5021,bojanowski2017enriching}.

{\em Word embeddings}, as these techniques are nowadays referred to, have been used extensively in SMM, both for topic detection \cite{sridhar2015unsupervised} and for sentiment analysis \cite{severyn2015twitter}.
%% MS: amaru, gräv fram lite referenser till användandet av embeddings för topic detection och sentiment analysis!
To the best of our knowledge, only one previous study \citep{dahlberg:2014} has used word embeddings to analyze issue ownership. 
%% MS: amaru: kolla att ovan stämmer!
However, that study relied on simple nearest neighbor analysis using cosine similarity to study language use in the Swedish blogosphere. 

We believe that prediction-based word embeddings such as Word2Vec are amenable to another notion of similarity, which we call predictive similarity. 
%% MS: and which is especially suited for measuring conditioned similarity... In order to demonstrate the viability of the predictive similarity measure, we utilize the measure to study issue ownership...

\subsection{Predictive Similarity}

Given a function $f : A \times B \rightarrow \mathbb{R}$, we define the predictive similarity of two items $x, y \in A$ as the correlation of $f(x, \mathbf{b})$, and $f(y, \mathbf{b})$, where $\mathbf{b}$ is a random variable of type $B$: 

\begin{equation}
\label{eq:psim}
	psim(x, y) = \frac{
    	\text{cov}\left(f\left(x, \mathbf{b}\right), f\left(y, \mathbf{b}\right)\right)
        }{
        \sqrt{\text{var}\left(f\left(x,\mathbf{b}\right)\right)\text{var}\left(f\left(y, \mathbf{b}\right)\right)}}
\end{equation}

At a very general level, prediction based word embeddings such as Word2Vec or FastText consists of a scoring function $s : C \times T \rightarrow \mathbb{R}$ with an objective function taking the following form: 
\begin{equation}
\sum_{t\times C \in D}\left[\sum_{c \in C} l(s(c, t)) + \sum_{n \in \mathcal{N}_{t,c}}l(-s(n, t)) \right]
\label{eq:cbow}
\end{equation}
where $l$ is the logistic loss function $l(x) = \log(1 + e^{-x})$ and $s$ being the model-specific scoring function that relates to the probability of observing the target $t$ in the context $c$. For the Skipgram variant of Word2Vec, this function $s$ is simply the dot product between a vector representation of the target word $t$, and a vector representation of the context word $c$.

The predictive similarity has several interpretations for the Skipgram model, but the simplest one is the one where we let $f = s$, i.e. we say that the similarity of two words $x$ and $y$ is the correlation between the scores they assign to target words $\mathbf{b}$, i.e. $\text{corr}(s(x, \mathbf{b}), s(y, \mathbf{b}))$. Since $s$ is linear, this correlation takes a fairly simple form:
\footnote{It might be interesting to note that this coincides with cosine similarity if $\text{var}(\mathbf{b})$ is a scalar multiple of the identity, i.e. if there is no correlation between dimensions and all dimensions have the same variance.}

\begin{equation}
\label{eq:cbow_psim}
\begin{split}
  & \text{cov} 
  (s(x, \mathbf{b}), f(s, \mathbf{b})) \\ &
  ~= 
  \mathbf{E}\left[
    \left(x^T \mathbf{b} - \overline{x^T\mathbf{b}}\right)
    \left(y^T \mathbf{b} - \overline{y^T\mathbf{b}}\right)
  \right]\\&
  ~= 
  \mathbf{E}\left[
    \left(x^T\left(\mathbf{b} - \overline{\mathbf{b}}\right)\right)
    \left(y^T\left(\mathbf{b} - \overline{\mathbf{b}}\right)\right)^T
  \right]\\&
  ~= 
  x^T\mathbf{E}\left[
    \left(\mathbf{b} - \overline{\mathbf{b}}\right)
    \left(\mathbf{b} - \overline{\mathbf{b}}\right)^T
  \right]y\\&
  ~= 
  x^T\text{var}(\mathbf{b})y\\&
  psim(x, y) = \frac{x^T\text{var}(\mathbf{b})y}
  {\sqrt{x^T\text{var}(\mathbf{b})x~y^T\text{var}(\mathbf{b})}y}
\end{split}
\end{equation}

We argue that we can get a a notion of conditioned similarity by estimating a \emph{weighted} correlation, where the weighting acts as the conditioning. 

\begin{table}
\begin{tabular}{lrrr}
\toprule 
\multicolumn{4}{c}{orange}\\ 
\hline
& \textbf{paint} & \textbf{juice} & \textbf{county} \\
\hline 
1 & deep-red & cranberry & siskiyou \\
2 & fuschia & lime & calaveras \\
3 & lime-green & caraway & ventura \\
4 & hand-woven & fanta & osceola \\
5 & blue & clove & yolo  \\ 
6 & yellow & zests & mendocino \\
7 & ocher & coconut & bernardino \\
8 & linoleum & peppercorns & okanogan \\ 
9 & duck-egg & lemons & okfuskee \\ 
10 & rust-colored & peach & tuolumne \\ 
\bottomrule 
\end{tabular}
\caption{Examples of predictive similarity neighborhoods of ``orange" conditioned on ``paint", ``juice", and ``county", respectively.   
\footnote{For this example, the underlying embedding is a 100-dimensional SkipGram model trained with negative sampling on the One Billion Word corpus \cite{chelba2013one}}
}
\label{tab:conditioning_example}
\end{table}

Table \ref{tab:conditioning_example} shows a small example where we  queried the neighborhood of the word ``orange", conditioned such that a single word (``paint'', ``juice'', and ``county'', respectively) accounts for half the weight in $\text{var}(\mathbf{b})$, with all other words in the vocabulary having equal weights.

Predictive similarity can easily be extended to similar models, and for the purpose of this paper in particular, we extend it to 
Doc2Vec \cite{le2014distributed}, a model where the notion of context is enriched by the source\footnote{By source we can mean a paragraph, document, or in our case: domain name from which the utterance originates.} of the utterance. The scoring function $s$ then takes the following form: $s(t, c, d) = t^T(c + d)$, with $d$ being a vector representation of the source in question.

We argue that by using conditioned predictive similarity on document embeddings we can answer questions such as: ``how similar is \emph{The BBC} to \emph{The Daily Mail}, when talking about \emph{Climate Change}". The end goal is to measure aggregate similarity in specific issues: ``when talking about \emph{health policy}, to which extent does the general language use align with \emph{Source A}, \emph{Source B}, \emph{Source C}, et.c.". 

\section{Experiments}

To answer the language similarity question posed by issue ownership we measure aggregate predictive similarity between party platforms and various subsets of online text data, conditioned on words pertaining to left wing issues, right wing issues, nativist issues, and general political topics.

\begin{table*}[t]
\begin{tabular}{lrrrr}
\toprule
Abbr. & Name & Translation & Word count & Bloc \\
\hline 
V & V\"ansterpartiet & The Left Party & 15,383 &  \multirow{3}{*}{Left} \\ 
S & Socialdemokraterna & The Social Democrats & 27,899 & \\ 
MP & Milj\"opartiet &  The Green Party & 19,471 & \\
\hline 
C & Centern & The Centre Party & 68,136 & \multirow{4}{*}{Right} \\
L & Liberalerna & The Liberals & 64,276 &  \\
KD & Kristdemokraterna & The Christian Democrats & 16,494 &  \\ 
M & Moderaterna & The Moderates & 12,807 & \\ 
\hline 
SD & Sverigedemokraterna & The Swedish Democrats & 3,430 & N/A (Nativist)\\
\hline 
FI & Feministiskt Initiativ & Feminist Initiative & 84,424 & N/A \\
\bottomrule
\end{tabular}
\caption{Party abbreviations, names, translated names, word count, and bloc allegiance.}
\label{tab:abbreviations}
\end{table*}

We built Doc2Vec embeddings \cite{le2014distributed} on Swedish online data from 2018 crawled by Trendiction and manually scraped  party platforms from the eight parties in parliament and \emph{Feministiskt Initiativ} (Feminist Initiative).\footnote{A complete list of parties, their abbreviations, their English translations, and bloc affiliation can be found in Table \ref{tab:abbreviations}.} Doc2Vec requires us to define a notion of source. For the data crawled by Trendiction, we take the source to be the domain name of the document, e.g.~\emph{www.wikipedia.se}, whereas for the manually scraped party platforms, we assign it the appropriate party identifier. The model was trained using the Gensim package \cite{rehurek_lrec} with embedding dimension 100 and a context window of size 8.

In collaboration with the Political Science department at Gothenburg University we also extracted keywords for each party from their party platform. We use these party specific keywords as a crude proxy for issues: we let left wing issues be defined by the union of left bloc party keywords, right wing issues be defined by right bloc party keywords, and nativist issues be defined by the keywords of Sverigedemokraterna (The Swedish Democrats), we also let the union of all keywords be representative for general political discourse. The parties' bloc alignment and the size of the data used to generate representations for them can be seen in Table \ref{tab:abbreviations}.

We let the conditioned predictive similarity between sources two $x$ and $y$ be defined by the following equation (Equation \ref{eq:doc_psim}), i.e.~a weighted variant of equation \ref{eq:cbow_psim}, where only words among the given issues keywords are accounted for, as described by Equation \ref{eq:weight}.

\begin{equation}
psim(x, y) = \frac{x^T\text{var}(\mathbf{t} ; w)y}
  {\sqrt{x^T\text{var}(\mathbf{t}; w)x~y^T\text{var}(\mathbf{t}; w)y}}
\label{eq:doc_psim}
\end{equation}
\begin{equation}
w_t = \begin{cases}
1, t \in \text{Issue keywords}\\
0, t \not \in \text{Issue keywords}
\end{cases} 
\label{eq:weight}
\end{equation}

Above, $x$ and $y$ are document vectors and $\text{var}(\mathbf{t}; w_t)$ is the weighted covariance matrix of the target word vectors. This is the equivalent of letting $s(d, c, t) = d^Tt$, i.e. the case we ignore the effect of context words.

Table \ref{tab:issueownership} (next side) shows the average predictive similarity between the political party platforms and various online data sources, conditioned on left wing party issues, right wing party issues, nativist party issues, and general political discourse. Average cosine similarity between the sources and parties is also shown as a comparison.

\begin{table*}
\begin{tabular}{llrrrrrrrrr}
\toprule
       &     &     V &     S &    MP &     C &     L &    KD &     M &    SD &    FI \\
Media & Issues &       &       &       &       &       &       &       &       &       \\
\midrule
\multirow{5}{*}{Left wing} & Left wing &  0.43 &  0.35 &  0.25 &  0.20 &  0.36 &  0.35 &  0.45 &  0.47 &  0.36 \\
       & Right wing &  0.44 &  0.38 &  0.36 &  0.34 &  0.41 &  0.36 &  0.45 &  0.45 &  0.32 \\
       & Nativist &  0.43 &  0.40 &  0.42 &  0.36 &  0.42 &  0.39 &  0.42 &  0.45 &  0.37 \\
       & All &  0.42 &  0.35 &  0.31 &  0.28 &  0.38 &  0.36 &  0.42 &  0.44 &  0.36 \\
\cline{2-11}
       & Cos &  0.50 &  0.48 &  0.48 &  0.46 &  0.51 &  0.47 &  0.53 &  0.49 &  0.44 \\
\cline{1-11}
\multirow{4}{*}{Right wing} & Left wing &  0.25 &  0.24 &  0.31 &  0.25 &  0.31 &  0.27 &  0.35 &  0.34 &  0.16 \\
       & Right wing &  0.28 &  0.31 &  0.32 &  0.32 &  0.34 &  0.28 &  0.36 &  0.36 &  0.19 \\
       & Nativist &  0.29 &  0.32 &  0.36 &  0.34 &  0.38 &  0.34 &  0.36 &  0.36 &  0.21 \\
       & All &  0.26 &  0.27 &  0.31 &  0.30 &  0.34 &  0.30 &  0.35 &  0.34 &  0.18 \\
\cline{2-11}
       & Cos & 0.44 &  0.45 &  0.44 &  0.47 &  0.51 &  0.47 &  0.51 &  0.45 &  0.41 \\
\cline{1-11}
\multirow{4}{*}{Nativist} & Left wing &  0.36 &  0.17 &  0.04 &  0.05 &  0.30 &  0.31 &  0.34 &  0.48 &  0.32 \\
       & Right wing &  0.28 &  0.09 &  0.08 &  0.17 &  0.30 &  0.32 &  0.30 &  0.39 &  0.23 \\
       & Nativist &  0.05 & -0.11 &  0.02 &  0.01 &  0.17 &  0.16 &  0.03 &  0.21 &  0.08 \\
       & All &  0.28 &  0.08 &  0.06 &  0.10 &  0.28 &  0.31 &  0.27 &  0.39 &  0.29 \\
\cline{2-11}
       & Cos &  0.51 &  0.45 &  0.47 &  0.45 &  0.56 &  0.53 &  0.56 &  0.61 &  0.53 \\
\cline{1-11}
\multirow{4}{*}{All News} & Left wing &  0.32 &  0.26 &  0.25 &  0.21 &  0.33 &  0.30 &  0.38 &  0.40 &  0.25 \\
       & Right wing &  0.33 &  0.30 &  0.30 &  0.31 &  0.36 &  0.32 &  0.38 &  0.40 &  0.24 \\
       & Nativist &  0.30 &  0.28 &  0.33 &  0.30 &  0.36 &  0.33 &  0.33 &  0.36 &  0.24 \\
       & All &  0.32 &  0.27 &  0.27 &  0.26 &  0.34 &  0.32 &  0.36 &  0.38 &  0.25 \\
\cline{2-11}
       & Cos &  0.47 &  0.46 &  0.46 &  0.47 &  0.52 &  0.48 &  0.52 &  0.48 &  0.44 \\
\cline{1-11}
\multirow{4}{*}{Social} & Left wing &  0.07 &  0.11 &  0.18 &  0.06 &  0.06 &  0.08 &  0.09 &  0.12 &  0.18 \\
       & Right wing &  0.20 &  0.28 &  0.31 &  0.22 &  0.18 &  0.14 &  0.20 &  0.17 &  0.26 \\
       & Nativist &  0.12 &  0.18 &  0.19 &  0.08 &  0.09 &  0.07 &  0.12 &  0.22 &  0.21 \\
       & All &  0.13 &  0.18 &  0.20 &  0.14 &  0.11 &  0.10 &  0.13 &  0.16 &  0.23 \\
\cline{2-11}
       & Cos &  0.42 &  0.42 &  0.42 &  0.40 &  0.39 &  0.41 &  0.42 &  0.45 &  0.39 \\
\bottomrule
\end{tabular}
\caption{Average predictive similarity (and cosine similarity) between political parties and various subsets of the online sources. 
}
\label{tab:issueownership}
\end{table*}

\section{Discussion}

\begin{table}[ht!]
\footnotesize
\begin{tabular}{llrrr}
\toprule
           & Bloc       &  Left &  Nativist &  Right \\
Media & Issues &       &           &        \\
\midrule
\multirow{3}{*}{Left wing} & Left wing & -0.02 &     -0.02 &  -0.06 \\
           & Nativist &  0.09 &     -0.00 &   0.03 \\
           & Right wing &  0.02 &     -0.05 &  -0.02 \\
\cline{1-5}
\multirow{3}{*}{Nativist} & Left wing &  0.02 &      0.18 &   0.04 \\
           & Nativist & -0.15 &     -0.06 &  -0.08 \\
           & Right wing & -0.03 &      0.08 &   0.05 \\
\cline{1-5}
\multirow{3}{*}{Right wing} & Left wing & -0.02 &     -0.06 &  -0.02 \\
           & Nativist &  0.07 &     -0.02 &   0.07 \\
           & Right wing &  0.01 &     -0.06 &  -0.01 \\
\bottomrule
\end{tabular}
\caption{
Grouped and normalized predictive similarity. 
}
\label{tab:normalized}
\end{table}

As can be seen in Table \ref{tab:issueownership}, there is a marked difference when conditioning on issues versus using regular document --- i.e. cosine --- similarity. Furthermore, we observe that conditioned similarity seems to align left wing media with left wing parties, nativist media with the Swedish Democrats, but not align right wing media with right wing parties.
This effect can be made more apparent by grouping the parties into blocs and fitting a simple additive model for the similarities along all dimensions (i.e.~Media, Issues, and Bloc), as a way to normalize for general Media, Issue, and Bloc similarity. The results of this normalization, i.e.~the residuals, can be observed in Table \ref{tab:normalized}. From this one can see a small trend where left wing media is similar to left wing parties, nativist media being similar to the Swedish Democrats, and both left wing media and right wing media being dissimilar to the Swedish Democrats.

Furthermore, we see a strong dissimilarity between nativist media and all parties regarding nativist issues. This is particularly true for parties promoting liberal immigration policy: The Left Party, The Social Democrats, The Green Party, The Centre Party, and The Moderates are all currently or historically promoting liberal immigration policy at odds with nativist sentiment. 

A shortcoming of the method used here is the rather limited amount of party specific data: the quality and the quantity of the text data used varies drastically between parties, as can be seen in Table \ref{tab:abbreviations}. Using, for example, parliamentary debates, opinion pieces, and other official party communication might improve data coverage.

\section{Conclusion}

In this paper we have introduced some very preliminary results on how to measure similarities in language use, conditioned on discourse, e.g. ``how similar is \emph{The BBC} to \emph{The Daily Mail}, when talking about \emph{Climate Change}". The end goal is to measure aggregate similarity in specific issues, answering questions such as ``when talking about \emph{health policy}, to which extent does the general language use align with \emph{Source A}, \emph{Source B}, etc.", and use such an aggregate measure to study issue ownership at scale.

We believe that issue ownership and agenda setting can be explored through the lens of language use and similarity, but deem it necessary to condition similarity to the specific issue at hand. The reason for this is the need to distinguish between level of engagement in an issue and agreement in an issue: two sources that talk a lot about an issue --- e.g.~health insurance --- but in very different ways should not be considered similar. Dually, if a source very rarely talks about an issue, but consistently does so in a way that is very similar to the way some political party talks about it, we consider it reasonable to believe that that source's opinion aligns with the political party in question on that specific issue.

While we have not found a satisfactory, direct, evaluation of this task, we do believe that the examples we put forward show some face validity of the proposed method at measuring ideological alignment. 

\section{Appendix}

\subsection{Left wing news sources}

\begin{compactitem}
\item Aftonbladet
\item Arbetarbladet
\item Dala-Demokraten
\item Folkbladet
\item ETC
\item Arbetaren
\item Flamman
\item Bang
\item Offensiv
\item Prolet\"aren
\end{compactitem}

\subsection{Right wing news sources}
\begin{compactitem}
\item Dagens Industri
\item Dalabygden
\item Hallands Nyheter
\item Axess
\item Svensk Tidskrift
\item Hemmets V\"an
\item Dagens Nyheter
\item G\"oteborgs-Posten
\item Helsingborgs Dagblad
\item Nerikes Allehanda
\item Sydsvenskan
\item Upsala Nya Tidning
\item Expressen
\item Svenska Dagbladet
\item Sm{\aa}landsposten
\item Norrbottens Kuriren
\end{compactitem}

\subsection{Nativist news sources}

\begin{compactitem}
\item Nordfront
\item Samh\"allsnytt
\item Fria Tider
\item Nya Tider
\item Samtiden
\end{compactitem}

\bibliographystyle{acl_natbib_nourl}
\bibliography{emnlp2018}

\begin{thebibliography}{29}
\expandafter\ifx\csname natexlab\endcsname\relax\def\natexlab#1{#1}\fi

\bibitem[{Allan et~al.(1998)Allan, Carbonell, Doddington, Yamron, Yang
  et~al.}]{allan1998topic}
James Allan, Jaime Carbonell, George Doddington, Jonathan Yamron, Yiming Yang,
  et~al. 1998.
\newblock Topic detection and tracking pilot study: Final report.
\newblock In \emph{Proceedings of the DARPA broadcast news transcription and
  understanding workshop}, volume 1998, pages 194--218. Citeseer.

\bibitem[{B\'{e}langer and Meguid(2008)}]{BELANGER2008477}
\'{E}ric B\'{e}langer and Bonnie~M. Meguid. 2008.
\newblock Issue salience, issue ownership, and issue-based vote choice.
\newblock \emph{Electoral Studies}, 27(3):477 -- 491.

\bibitem[{Bermingham and Smeaton(2011)}]{bermingham2011using}
Adam Bermingham and Alan Smeaton. 2011.
\newblock On using twitter to monitor political sentiment and predict election
  results.
\newblock In \emph{Proceedings of the Workshop on Sentiment Analysis where AI
  meets Psychology (SAAIP 2011)}, pages 2--10.

\bibitem[{Bojanowski et~al.(2017)Bojanowski, Grave, Joulin, and
  Mikolov}]{bojanowski2017enriching}
Piotr Bojanowski, Edouard Grave, Armand Joulin, and Tomas Mikolov. 2017.
\newblock Enriching word vectors with subword information.
\newblock \emph{Transactions of the Association for Computational Linguistics},
  5:135--146.

\bibitem[{Bullinaria and Levy(2007)}]{Bullinaria2007}
John~A. Bullinaria and Joseph~P. Levy. 2007.
\newblock Extracting semantic representations from word co-occurrence
  statistics: A computational study.
\newblock \emph{Behavior Research Methods}, 39(3):510--526.

\bibitem[{Ceron et~al.(2015)Ceron, Curini, and Iacus}]{ceron2015using}
Andrea Ceron, Luigi Curini, and Stefano~M Iacus. 2015.
\newblock Using sentiment analysis to monitor electoral campaigns: Method
  matters—evidence from the united states and italy.
\newblock \emph{Social Science Computer Review}, 33(1):3--20.

\bibitem[{Chelba et~al.(2013)Chelba, Mikolov, Schuster, Ge, Brants, Koehn, and
  Robinson}]{chelba2013one}
Ciprian Chelba, Tomas Mikolov, Mike Schuster, Qi~Ge, Thorsten Brants, Phillipp
  Koehn, and Tony Robinson. 2013.
\newblock One billion word benchmark for measuring progress in statistical
  language modeling.
\newblock \emph{arXiv preprint arXiv:1312.3005}.

\bibitem[{Dahlberg and Sahlgren(2014)}]{dahlberg:2014}
Stefan Dahlberg and Magnus Sahlgren. 2014.
\newblock Issue framing and language use in the swedish blogosphere: Changing
  notions of the outsider concept.
\newblock In Bertie Kaal, Isa Maks, and Annemarie van Elfrinkhof, editors,
  \emph{From Text to Political Positions: Text Analysis across Disciplines},
  pages 71--92. John Benjamins.

\bibitem[{Deerwester et~al.(1990)Deerwester, Dumais, Furnas, Landauer, and
  Harshman}]{deerwester}
Scott Deerwester, Susan~T. Dumais, George~W. Furnas, Thomas~K. Landauer, and
  Richard Harshman. 1990.
\newblock Indexing by latent semantic analysis.
\newblock \emph{Journal of the American Society for Information Science},
  41(6):391--407.

\bibitem[{Kiousis et~al.(2015)Kiousis, Str\"{o}mb\"{a}ck, and
  McDevitt}]{IJoC3008}
Spiro Kiousis, Jesper Str\"{o}mb\"{a}ck, and Michael McDevitt. 2015.
\newblock Influence of issue decision salience on vote choice: Linking agenda
  setting, priming, and issue ownership.
\newblock \emph{International Journal of Communication}, 9(0).

\bibitem[{Kl\"{u}ver and {n}aki Sagarzazu(2016)}]{kluver}
Heike Kl\"{u}ver and I\~{n}aki Sagarzazu. 2016.
\newblock Setting the agenda or responding to voters? political parties, voters
  and issue attention.
\newblock \emph{West European Politics}, 39(2):380--398.

\bibitem[{{Landauer} and {Dumais}(1997)}]{landauer_solution_1997}
{Thomas K} {Landauer} and {Susan T} {Dumais}. 1997.
\newblock A solution to plato's problem: The latent semantic analysis theory of
  acquisition, induction, and representation of knowledge.
\newblock \emph{Psychological review}, 104(2):211--240.

\bibitem[{Le and Mikolov(2014)}]{le2014distributed}
Quoc Le and Tomas Mikolov. 2014.
\newblock Distributed representations of sentences and documents.
\newblock In \emph{International Conference on Machine Learning}, pages
  1188--1196.

\bibitem[{Liu(2012)}]{Liu:2012:SAO:3019323}
Bing Liu. 2012.
\newblock \emph{Sentiment Analysis and Opinion Mining}.
\newblock Morgan \& Claypool Publishers.

\bibitem[{Mccombs and Reynolds(2002)}]{mccombs}
Maxwell Mccombs and Amy Reynolds. 2002.
\newblock News influence on our pictures of the world.
\newblock In Jennings Bryant and Dolf Zillmann, editors, \emph{Media Effects.
  Advances in Theory and Research}, pages 1--18. Lawrence Erlbaum Associates.

\bibitem[{Mikolov et~al.(2013)Mikolov, Sutskever, Chen, Corrado, and
  Dean}]{NIPS2013_5021}
Tomas Mikolov, Ilya Sutskever, Kai Chen, Greg~S Corrado, and Jeff Dean. 2013.
\newblock Distributed representations of words and phrases and their
  compositionality.
\newblock In C.~J.~C. Burges, L.~Bottou, M.~Welling, Z.~Ghahramani, and K.~Q.
  Weinberger, editors, \emph{Advances in Neural Information Processing Systems
  26}, pages 3111--3119. Curran Associates, Inc.

\bibitem[{Pang and Lee(2008)}]{Pang:2008:OMS:1454711.1454712}
Bo~Pang and Lillian Lee. 2008.
\newblock Opinion mining and sentiment analysis.
\newblock \emph{Foundations and Trends in Information Retrieval},
  2(1-2):1--135.

\bibitem[{Pennington et~al.(2014)Pennington, Socher, and
  Manning}]{pennington2014glove}
Jeffrey Pennington, Richard Socher, and Christopher~D Manning. 2014.
\newblock Glove: Global vectors for word representation.
\newblock In \emph{EMNLP}, volume~14, pages 1532--1543.

\bibitem[{Pozzi et~al.(2016)Pozzi, Fersini, Messina, and
  Liu}]{Pozzi:2016:SAS:3044756}
Federico~Alberto Pozzi, Elisabetta Fersini, Enza Messina, and Bing Liu. 2016.
\newblock \emph{Sentiment Analysis in Social Networks}, 1st edition.
\newblock Morgan Kaufmann Publishers Inc., San Francisco, CA, USA.

\bibitem[{{\v R}eh{\r u}{\v r}ek and Sojka(2010)}]{rehurek_lrec}
Radim {\v R}eh{\r u}{\v r}ek and Petr Sojka. 2010.
\newblock {Software Framework for Topic Modelling with Large Corpora}.
\newblock In \emph{{Proceedings of the LREC 2010 Workshop on New Challenges for
  NLP Frameworks}}, pages 45--50, Valletta, Malta. ELRA.
\newblock \url{http://is.muni.cz/publication/884893/en}.

\bibitem[{Sahlgren(2006)}]{sahlgren2006word}
Magnus Sahlgren. 2006.
\newblock \emph{{The Word-space model}}.
\newblock Ph.D. thesis, University of Stockholm (Sweden).

\bibitem[{Salton(1971)}]{Salton:1971:SRS:1102022}
Gerard Salton. 1971.
\newblock \emph{The SMART Retrieval System---Experiments in Automatic Document
  Processing}.
\newblock Prentice-Hall, Inc., Upper Saddle River, NJ, USA.

\bibitem[{Sch\"{u}tze(1992)}]{Schutze:1992:DM:147877.148132}
Hinrich Sch\"{u}tze. 1992.
\newblock Dimensions of meaning.
\newblock In \emph{Proceedings of the 1992 ACM/IEEE Conference on
  Supercomputing}, Supercomputing '92, pages 787--796, Los Alamitos, CA, USA.
  IEEE Computer Society Press.

\bibitem[{Sch\"{u}tze(1993)}]{Schütze93wordspace}
Hinrich Sch\"{u}tze. 1993.
\newblock Word space.
\newblock In \emph{Advances in Neural Information Processing Systems 5}, pages
  895--902. Morgan Kaufmann.

\bibitem[{Severyn and Moschitti(2015)}]{severyn2015twitter}
Aliaksei Severyn and Alessandro Moschitti. 2015.
\newblock Twitter sentiment analysis with deep convolutional neural networks.
\newblock In \emph{Proceedings of the 38th International ACM SIGIR Conference
  on Research and Development in Information Retrieval}, pages 959--962. ACM.

\bibitem[{Sridhar(2015)}]{sridhar2015unsupervised}
Vivek Kumar~Rangarajan Sridhar. 2015.
\newblock Unsupervised topic modeling for short texts using distributed
  representations of words.
\newblock In \emph{Proceedings of the 1st workshop on vector space modeling for
  natural language processing}, pages 192--200.

\bibitem[{Stubager(2018)}]{Stubager2018}
Rune Stubager. 2018.
\newblock What is issue ownership and how should we measure it?
\newblock \emph{Political Behavior}, 40(2):345--370.

\bibitem[{Turney(2002)}]{Turney:2002:TUT:1073083.1073153}
Peter~D. Turney. 2002.
\newblock Thumbs up or thumbs down?: Semantic orientation applied to
  unsupervised classification of reviews.
\newblock In \emph{Proceedings of the 40th Annual Meeting on Association for
  Computational Linguistics}, ACL '02, pages 417--424, Stroudsburg, PA, USA.
  Association for Computational Linguistics.

\bibitem[{Turney and Pantel(2010)}]{Turney:2010:FMV:1861751.1861756}
Peter~D. Turney and Patrick Pantel. 2010.
\newblock From frequency to meaning: Vector space models of semantics.
\newblock \emph{Journal of Artificial Intelligence Research}, 37(1):141--188.

\end{thebibliography}

\end{document}